\documentclass[wcp]{jmlr}

\usepackage{longtable}

\usepackage{booktabs}

\usepackage{amsfonts}       
\usepackage{nicefrac}       
\usepackage{microtype}      
\usepackage{graphicx}
\usepackage{enumitem}
\usepackage{tabu}
\usepackage{algpseudocode}
\usepackage{mathtools}
\usepackage{amsmath}
\usepackage{array}
\usepackage{fancyhdr}
\usepackage{natbib}
\usepackage{amssymb}
\usepackage{tikz}
\usepackage{bussproofs}
\usepackage[utf8]{inputenc}
\usepackage{multirow}
\usepackage{color}
\usepackage{makecell}
\usepackage{arydshln}
\usepackage{algorithm}

\usepackage{wrapfig}

\newcommand{\ourl}{Sketched Leverage Score Ordering}
\newcommand{\ours}{SLSO}

\jmlrvolume{80}
\jmlryear{2018}
\jmlrworkshop{ACML 2018}

\title[An Empirical Evaluation of Sketched SVD and Leverage Score Ordering]{An Empirical Evaluation of Sketched SVD and its Application to Leverage Score Ordering}

  \author{\Name{Hui Han Chin} \Email{hhchin87@gmail.com}\\
  \addr School of Computer Science, Carnegie Mellon University
  \AND
  \Name{Paul Pu Liang} \Email{pliang@cs.cmu.edu}\\
  \addr Machine Learning Department, Carnegie Mellon University
 }

\begin{document}

\maketitle

\begin{abstract}
The power of randomized algorithms in numerical methods have led to fast solutions which use the Singular Value Decomposition (SVD) as a core routine. However, given the large data size of modern and the modest runtime of SVD, most practical algorithms would require some form of approximation, such as sketching, when running SVD. While these approximation methods satisfy many theoretical guarantees, we provide the first algorithmic implementations for sketch-and-solve SVD problems on real-world, large-scale datasets. We provide a comprehensive empirical evaluation of these algorithms and provide guidelines on how to ensure accurate deployment to real-world data. As an application of sketched SVD, we present Sketched Leverage Score Ordering, a technique for determining the ordering of data in the training of neural networks. Our technique is based on the distributed computation of leverage scores using random projections. These computed leverage scores provide a flexible and efficient method to determine the optimal ordering of training data without manual intervention or annotations. We present empirical results on an extensive set of experiments across image classification, language sentiment analysis, and multi-modal sentiment analysis. Our method is faster compared to standard randomized projection algorithms and shows improvements in convergence and results.
\end{abstract}


\section{Introduction}

Many problems in numerical methods rely heavily on the Singular Value Decomposition (SVD), a technique for factorizing a real or complex matrix. Such problems include low-rank approximations, computation of leverage scores, pseudo-inverses, and including their weighted and robust variants. These applications have become more important with the advent of big data and large-scale machine learning. For example, weighted low rank approximations are used in recommendation systems~\citep{Razenshteyn:2016:WLR:2897518.2897639,Markovsky:2008:SLA:1354944.1355138} and leverage scores are heavily used in statistical data analysis~\citep{10.2307/4616482,DBLP:journals/corr/drineas11}. While the exact computation of the SVD can provide deterministic solutions to these problems, these algorithms are often too slow on large-scale datasets. As a result, research has focused on efficient approximate algorithms for scalable computation of the SVD.

Recently, the power of randomized algorithms has led to much faster solutions to these problems. These randomized algorithms for numerical linear algebra were popularized by the sketching method~\citep{Woodruff:2014:STN:2693651.2693652}. The basis for these sketching algorithms involves multiplying the data matrix $\mathbf{X} \in \mathbb{R}^{n \times d}$ with a smaller randomized matrix. The sketched problem is then smaller and can be solved using the same algorithm in faster input runtime. Although the truncated SVD~\citep{Hansen1987} and other iterative SVD algorithms~\citep{6252789} also provide efficient alternatives, sketching algorithms are more efficient and accurate, especially when the input is sparse~\citep{Woodruff:2014:STN:2693651.2693652}.

While these sketching methods satisfy theoretical guarantees~\citep{DBLP:journals/corr/drineas11,Woodruff:2014:STN:2693651.2693652} (see details in section~\ref{s2}), our contribution lies in the first implementations of these algorithms for sketching SVD problems on real-world, large-scale datasets. We found that the naive implementation of sketched SVD leads to several problems such as difficulties in estimating true ranks of real-world datasets, corruption by high-dimensional noise, and impact of small singular values. We provide a comprehensive empirical evaluation of these algorithms and summarize guidelines on how to ensure accurate deployment to real-world data. Finally, we examine an application of sketched SVD on the efficient computation of leverage scores. We apply the sketched leverage scores as an ordering method for curriculum learning, and show improved performance over random ordering when training deep neural networks on three datasets. Although~\citep{DBLP:conf/kdd/DahiyaKW18} also empirically evaluate sketching, they focus on general algorithms rather than a treatment of SVD problems. We further provide an application of sketched SVD for curriculum learning.

The remainder of the paper is organized as follows. We first state several preliminary results in section~\ref{s2}. We then provide valuable insights into the practical implementations of sketched SVD on large-scale datasets in section~\ref{s3}. In section~\ref{s4} we outline the application of sketched SVD: the \ourl \ algorithm. We present experimental results in section~\ref{s5}. We then discuss our observations in section~\ref{s6}. Finally, we describe related work and conclude the paper in sections~\ref{s7} and~\ref{s8} respectively.
\section{Preliminaries}
\label{s2}
\subsection{Sketching and Subspace Embeddings}

Popular choices for the randomized sketching matrix include random Gaussian matrices~\citep{Woodruff:2014:STN:2693651.2693652}, the Fast Johnson Lindenstrauss transform (FJLT)~\citep{Sarlos:2006:IAA:1170136.1170496}, the Subsampled Randomized Hadamard Transform (SRHT)~\citep{DBLP:journals/corr/abs-1204-0062,journals/aada/Tropp11}, the CountSketch matrix~\citep{Bourgain:2015:TUT:2746539.2746541,Clarkson:2017:LAR:3038256.3019134,DBLP:conf/soda/Cohen16,Meng:2013:LSE:2488608.2488621} or the OSNAP matrix~\citep{Nelson:2013:OFN:2570450.2570582}. All these sketching matrix $\mathbf{S}$ are useful because they satisfy the subspace embedding property:

\begin{definition}{Subspace Embedding: }
Let $\mathbf{A}$ be a $n$ by $d$ matrix. A $(1\pm \epsilon)$ $\ell$2 \textit{subspace embedding} for the column space of $\mathbf{A}$ is a k by n matrix $\mathbf{S}$ such that $\forall x \in \mathbb{R}^n$,
$$ (1-\epsilon) \|\mathbf{A}x\|^2_2 \leq \|\mathbf{SA}x\|^2_2 \leq (1 + \epsilon)\|\mathbf{A}x\|^2_2$$
\end{definition}

\begin{theorem}{}
\label{thr}
A subspace embedding preserves the set of singular values, of the input matrix $\mathbf{A}$. In particular, if $\mathbf{S}$ is a $(1\pm \epsilon)$ $\ell$2 subspace embedding for $\mathbf{A}$, then:
$$ \sigma_k (\mathbf{SA}) = (1 \pm O(\epsilon)) \sigma_k (\mathbf{A}) $$
\end{theorem}

A proof of theorem~\ref{thr} is given in~\citep{Li:2014:SMN:2634074.2634188}. In addition to least squares regression, L1 regression, and low-rank approximations~\citep{Woodruff:2014:STN:2693651.2693652}, more recent applications of sketching also include low-rank tensor regression~\citep{DBLP:journals/corr/abs-1709-07093}, $q$ to $p$ norms~\citep{DBLP:journals/corr/abs-1806-06429}, and Kronecker product regression~\citep{DBLP:conf/aistats/DiaoSSW18}.

\subsection{Leverage Scores Sampling}
The subspace embedding methods presented above are oblivious to the structure of the input matrix $\mathbf{A}$. For example if a \texttt{CountSketch} matrix $\mathbf{S}$ is used to embed $\mathbf{A}$ and $\mathbf{A}$ has sparse rows, then $\mathbf{SA}$ has sparse rows too. However, there are applications in which the goal is to sample the important rows thus having a sparse $\mathbf{SA}$ is not helpful. 

The leverage score of a data point is a measure of how much an outlier the data point is from other points in data $\mathbf{A}$. The underlying data model is assumed to be linear and the leverage score of the $i$th point, $l_i$, is given by $i$th row norm of the projection matrix of $\mathbf{A}$, i.e
$$l_i = |[\mathbf{H}]_i|^2_2, \; \mathbf{H} = \mathbf{A}(\mathbf{A}^T\mathbf{A})^+\mathbf{A}^T$$
Alternatively, taking the Singular Value Decomposition, $\mathbf{A} = \mathbf{U\Sigma V}^T$, $l_i$ can be expressed as
$$l_i = |\mathbf{U}_{i,*}|^2_2$$
Computing the leverage scores naively requires a costly matrix inversion, which can be prohibitive for very large data sets. Fortunately, the techniques of matrix sketching can be used to approximate the leverage scores~\citep{DBLP:journals/corr/drineas11}.

\section{Practical Implementation of Fast Leverage Scores}
\label{s3}

\citep{DBLP:journals/corr/drineas11} showed that sketching produces a small but good subspace embedding for finding the orthonormal basis. In the sketch space, the orthonormal basis can then computed using a Singular Value Decomposition (SVD). Also, \citep{DBLP:journals/corr/drineas11} gave the procedures of finding 1) exact leverage scores by SVD and 2) an $(1+\epsilon)$ approximation to the leverage scores by sketching:

\noindent \begin{algorithm}[htbp]
\SetAlgoLined
\SetKwInOut{Input}{Input}\SetKwInOut{Output}{Output}
\caption{Exact Leverage Score by SVD}
\Input{Given $n\times d$ matrix $\mathbf{A}$.}
\Output{Leverage score of $i$th row as $l_i$.}
1. Compute SVD, $\mathbf{A} = \mathbf{U\Sigma V}^T$.

2. Compute $l_i$ from the first $d$ columns of the $i$th row $l_i= |\mathbf{U}_i|^2_2$.
\label{a1}
\end{algorithm}

\noindent \begin{algorithm}[H]
\SetAlgoLined
\SetKwInOut{Input}{Input}\SetKwInOut{Output}{Output}
\caption{Approximation Leverage Score by Sketching}
\Input{Given $n\times d$ matrix $\mathbf{A}$.}
\Output{Approximate Leverage score of $i$th row as $l_i$.}
1. Compute Sketch of $\mathbf{A}$, $\mathbf{SA}$.

2. Compute SVD, $\mathbf{SA} = \mathbf{U\Sigma V}^T$.

3. Compute $\mathbf{U}^{approx} = \mathbf{AV}^T\mathbf{\Sigma}^{-1}$.

4. Compute $l_i$ from the first $D$ columns of the $i$th row $l_i= |\mathbf{U}^{approx}_i|^2_2$.
\label{a2}
\end{algorithm}

\subsection{Implementation Details}
The following sketching schemes were implemented in Python 3.6 and tested for their performance on a 64-bit machines with i7 CPU with 12 cores and 32 GB RAM.

\subsubsection{SVD for Exact Leverage Scores}
The SVD used for the exact computation of leverage scores (Algorithm~\ref{a1}) is from numpy.linalg.svd which uses LAPACK's SVD routines. It is highly optimized and multithreaded, take advantaging of the multi-cores CPU.

\subsubsection{Approximate Leverage Scores}
To compute approximate leverage scores via sketching (Algorithm~\ref{a2}), we examine the following choices for sketching matrices $\mathbf{S}$: \\

\noindent \textbf{Subsampled Randomized Hadamard Transform}: A Fast Johnson Lindenstrauss sketching matrix was implemented as a Subsampled Randomized Hadamard Transform (SRHT). The core Hadamard Transform was implemented using the Faster Fast Hadamard Transform (FFHT) of \citep{DBLP:journals/corr/AndoniILRS15} and uses Advanced Vector Extensions (AVX) CPU instructions to speed up the Hadamard Transform. Below is a synthetic data test run for comparing SVD vs SRHT to compute leverage scores (Table~\ref{t1}). However, we decided to drop SRHT from future experiments as it is difficult to meet the SRHT unique requirement of the row having to be a power of 2.

\begin{table}[!htb]
\fontsize{10}{10}\selectfont
\centering
\setlength\extrarowheight{2pt}
\setlength\tabcolsep{9.35pt}
\begin{tabular}{l : c c c c c}
\Xhline{3\arrayrulewidth}
Data size = $n \times$ 50 & \multicolumn{5}{c}{Running Time in Seconds} \\
$n =$      & $2^{16}$ & $2^{18}$ & $2^{20}$ & $2^{22}$ & $2^{24}$  \\ 
\Xhline{0.5\arrayrulewidth}
Sketching & 0.9 & 3.3 & 5.4 & 13.5 & 122.5 \\
SVD & 0.7 & 3.3 & 13.9 & 57.2 & 2114.0 \\
\Xhline{3\arrayrulewidth}
\end{tabular}
\caption{Running time for Synthetic Data. Sketching's $\epsilon=0.25$. Sketching matrices indeed increase running time significantly when $n$ increases.}
\label{t1}
\end{table}
\noindent \textbf{{CountSketch}, {OSNAP}}: OSNAP is a variant of CountSketch as described in \citep{Nelson:2013:OFN:2570450.2570582}. Both CountSketch and OSNAP are implemented in Python using the Numpy Library without additional acceleration. They are implemented in the streaming model where the sketching matrix rows are updated as the rows of the original matrix are read. Both sketches are implemented as a pair of hash functions.

\subsubsection{Distributed Sketch, Coordinator Model}
We implemented the distributed sketching, coordinator model framework for CountSketch and OSNAP as described in \citep{DBLP:journals/corr/BalcanLSW015}. The simulation of a distributed computing network was done using multiple processes using the Multiprocessing library in python. To simulate a row partition model, the data was fed to each process before sketching was completed. A single process acted as the coordinator. In this setup, there is minimal overhead penalty for the communication cost. 

\subsection{Implementation Experiments for Timing Performance}
The following experiments were meant to compare the performance of approximate leverage scores under different matrices. The number of rows for each sketching matrix was determined using their theoretical guarantees: $O((d/\epsilon)^2)$ for CountSketch and $O(d/\epsilon^2\log d)$ for OSNAP. For all experiments, the timings shown are for the actual time taken to compute the leverage scores. This excludes the time for data generation, setup and the output of the final scores. 

\noindent \textbf{Synthetic Data, Skinny Matrix:} Table~\ref{t2} shows that CountSketch is competitive when the number of column dimensions are small.
\begin{table}[!htb]
\fontsize{10}{10}\selectfont
\centering
\setlength\extrarowheight{4pt}
\setlength\tabcolsep{9.35pt}
\begin{tabular}{l : c c c}
\Xhline{3\arrayrulewidth}
K & \textbf{SVD} & \textbf{CountSketch} & \textbf{OSNAP} \\
\Xhline{0.5\arrayrulewidth}
10 & 0.001 & 0.22 & 0.22 \\ 
14 & 0.009 & 0.23 & 0.63 \\ 
18 & 0.21  & 0.73 & 5.12 \\ 
22 & 3.76  & 0.11 & 73.5 \\
\Xhline{3\arrayrulewidth}
\end{tabular}
\caption{$N=2^k, D=16$, 8 Threads, Sketching's $\epsilon=0.5$, Timings in seconds.}
\label{t2}
\end{table}

\noindent \textbf{Synthetic Data:} Table~\ref{t3} shows that sketching is faster than SVD as the matrices increase in size.
\begin{table}[!htb]
\fontsize{10}{10}\selectfont
\centering
\setlength\extrarowheight{4pt}
\setlength\tabcolsep{9.35pt}
\begin{tabular}{l : c c c}
\Xhline{3\arrayrulewidth}
K & \textbf{SVD} & \textbf{CountSketch} & \textbf{OSNAP} \\
\Xhline{0.5\arrayrulewidth}
10 & 0.06 & 0.28 & 0.29 \\ 
14 & 0.41 & 1.02 & 0.98 \\ 
18 & 10.26  & 16.44 & 12.15 \\ 
22 & 257.55  & 45.16& 169.44 \\
\Xhline{3\arrayrulewidth}
\end{tabular}
\caption{$N=2^k, D=256$, 8 Threads, Sketching's $\epsilon=0.5$, Timings in seconds.}
\label{t3}
\end{table}

\noindent \textbf{Synthetic Data, Squarish Matrix:} Table~\ref{t4} shows that OSNAP sketching is slightly faster than SVD for large matrices in both rows and columns. CountSketch was not used as it does not meet the requirements on the number of rows.
\begin{table}[!htb]
\fontsize{10}{10}\selectfont
\centering
\setlength\tabcolsep{9.35pt}
\setlength\extrarowheight{4pt}
\begin{tabular}{l : c c}

\Xhline{3\arrayrulewidth}
K & \textbf{SVD} & \textbf{OSNAP} \\
\Xhline{0.5\arrayrulewidth}
16 & 192.43 & 155.53 \\ 
\Xhline{3\arrayrulewidth}
\end{tabular}
\caption{$N=2^{16}, D=2^{12}$ 8 Threads, Sketching's $\epsilon=0.5$, Timings in seconds.}
\label{t4}
\end{table}

\subsection{Implementation Experiments on the Effects of Column Rank on Leverage Scores Approximation}
An assumption in most of the bounds for sketching matrices for subspace embedding is that the matrix $A$ has full column rank~\citep{DBLP:journals/corr/drineas11,Clarkson:2017:LAR:3038256.3019134}. However this is not a realistic assumption for most real world data as the latent dimension of the data might not be known. Real world data is often generated from a low dimensional subspace and corrupted with high dimensional noise. The following experiments test the effects of getting a wrong estimate for the column rank of the matrix that sketching is applied on. The two sketching matrices tested were CountSketch and SRHT. OSNAP was not tested as it can be seen as a variant of CountSketch.
\begin{figure}[H]
\centering
\begin{tabular}{cc}
\includegraphics[scale=0.4]{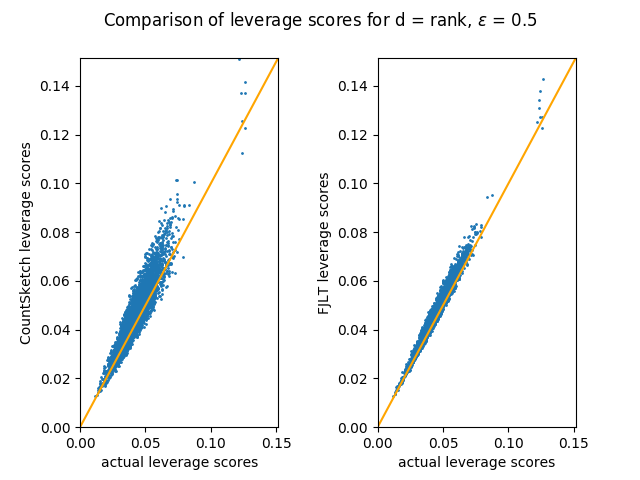} &
\includegraphics[scale=0.4]{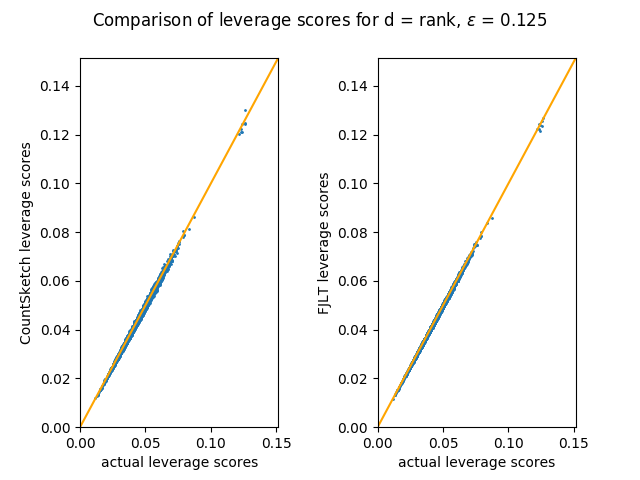} 
\end{tabular}
\caption{Plot of sketched (Countsketch, FJLT) vs actual leverage scores when the data matrix has full column rank. Approximate leverage scores are close to the actual scores.}
\label{f1}
\end{figure}
\vspace{-4mm}

\begin{figure}[H]
\centering
\begin{tabular}{cc}
\includegraphics[scale=0.4]{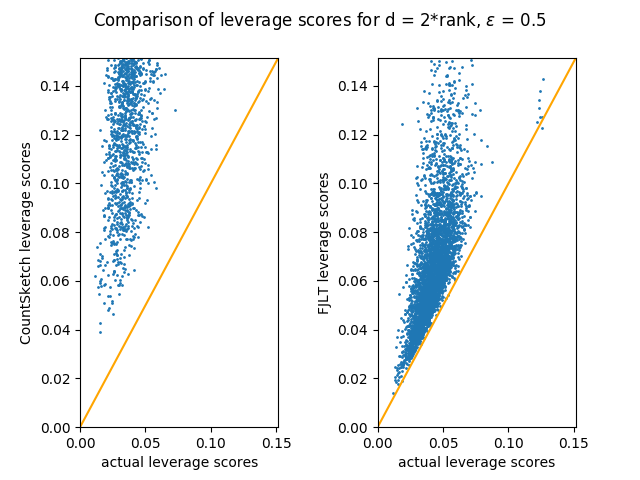} &
\includegraphics[scale=0.4]{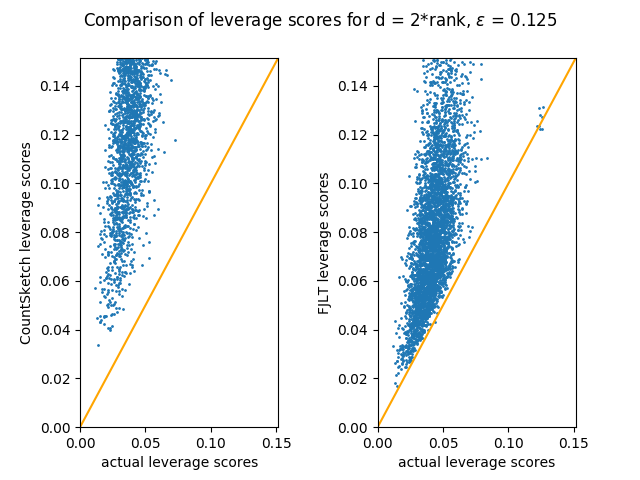} 
\end{tabular}
\caption{Plot of sketched (Countsketch, FJLT) vs actual leverage scores when the data matrix has column rank half of column dimension. When the column rank is wrong, even by a factor of 2, the approximate leverage scores computed can be significantly off.}
\label{f2}
\end{figure}
\vspace{-4mm}

From Figures~\ref{f1} and~\ref{f2}, we observe that when the column rank is the same as the dimension, then as suggested by the theoretical bounds, the quality of the approximated leverage scores is a function of the approximation factor $\epsilon$. However, when the column rank is wrong, even by a factor of 2, then the approximate leverage scores computed can be significantly off and the approximation error can be unbounded.

\begin{figure}[H]
\centering
\includegraphics[scale=0.5]{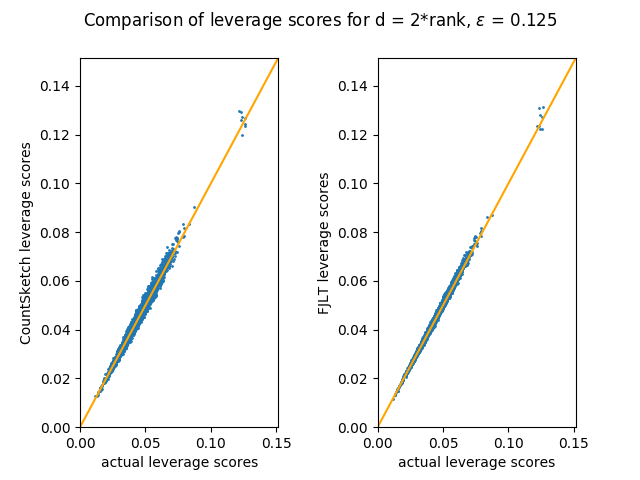}
\caption{Plot of sketched (Countsketch, FJLT) vs actual leverage scores when the  column orthonormal basis that is computed from the sketched matrix is truncated to the true rank of the original data. Approximate leverage scores are now close to the actual scores, as desired.}
\label{f3}
\end{figure}
However, if the column orthonormal basis that is computed from the sketched matrix is truncated to the true rank of the original data, the approximate leverage score is a function of the approximation factor $\epsilon$ (Figure~\ref{f3}). Both experiments show that having the column rank meeting the column dimension is a hard requirement, failing which would result in an unbounded approximation error.

\subsection{Effects of Small Singular Values on Leverage Scores Approximation}
\begin{figure}[h]
\centering
\includegraphics[scale=0.35]{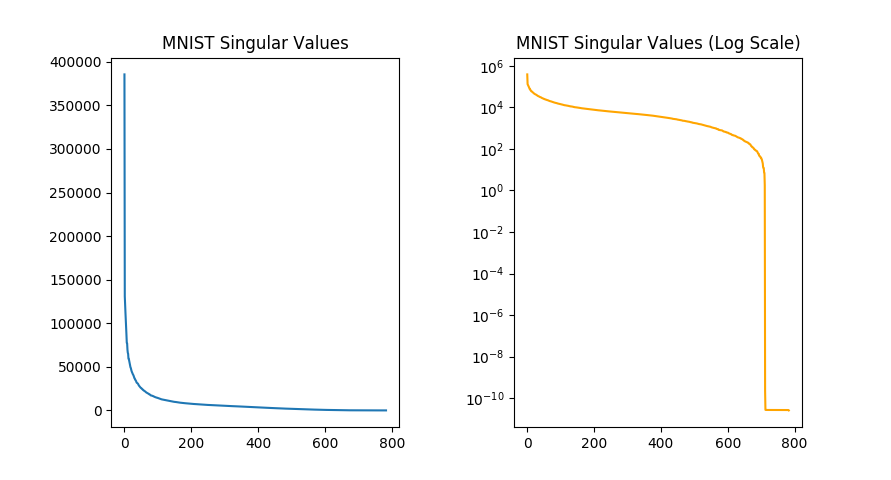}
\caption{Singular values of the MNIST dataset, on original (left) and log (right) scale.}
\label{f4}
\end{figure}

When the data is corrupted by small amount of high rank noise, it would appear as though it has full column rank. An example of high rank noise would be additive Gaussian white noise. For example, shown in Figure~\ref{f4} is a plot of the singular values of the MNIST handwriting training data, where each training example is flattened into a row vector and the entire data set is stacked vertically to form a $60000 \times 764$ matrix.

After applying the approximation leverage score described so far, we can see from Figure~\ref{f5} (left) that the approximation error can be much larger than the $\epsilon$ factor used and it looks similar to the case where the column rank was less than the column dimension. The error can be attributed to numerical precision errors when very small singulars are inverted  to compute the orthonormal basis from the sketch. i.e step 3, $\mathbf{U}^{\text{approx}} = \mathbf{AV}^T\mathbf{\Sigma}^{-1}$.

\subsection{Modified Leverage Scores Approximation by Sketching}

\noindent \begin{algorithm}[H]
\SetAlgoLined
\SetKwInOut{Input}{Input}\SetKwInOut{Output}{Output}
\caption{Approximate Leverage Score with Truncation}
\Input{Given $n\times d$ matrix $\mathbf{A}$, threshold $\epsilon$.}
\Output{Approximate Leverage score of $i$th row as $l_i$.}
1. Compute Sketch of $\mathbf{A}$, $\mathbf{SA}$.

2. Compute SVD, $\mathbf{SA = U\Sigma V}^T$.  

3. Truncate $\mathbf{V}, \mathbf{\Sigma}$ at small singular values less than $\epsilon$. Let this truncated matrices be $\mathbf{V}'$, $\mathbf{\Sigma'}$

4. Compute $\mathbf{U}^{approx} = \mathbf{AV'}^T\mathbf{\Sigma'}^{-1}$.

5. Compute $l_i$ from the first $d$ columns of the $i$th row $l_i= |\mathbf{U}^{approx}_i|^2_2$.
\label{a3}
\end{algorithm}

\begin{figure}[H]
\centering
\begin{tabular}{cc}
\includegraphics[scale=0.4]{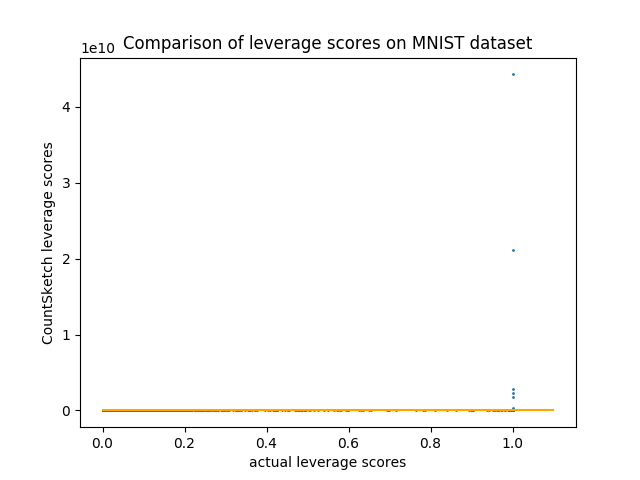} &
\includegraphics[scale=0.4]{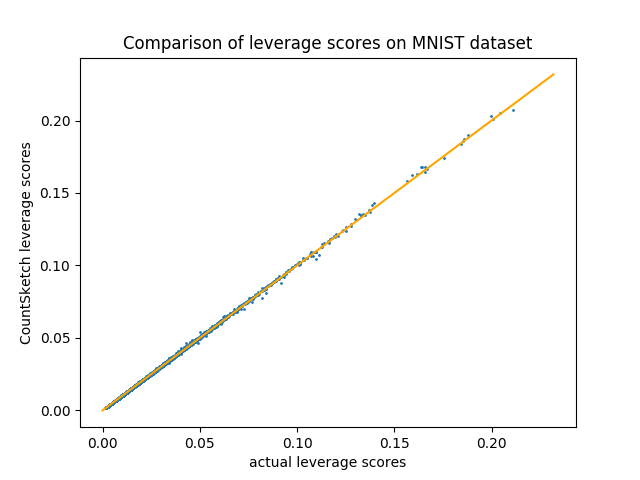} 
\end{tabular}
\caption{Sketched leverage scores in comparison with actual leverage scores. Left: without truncation, right: after truncation for smallest singular values, giving the desired accuracies.}
\label{f5}
\end{figure}

Small singular values would appear in the sketched matrix $\mathbf{SA}$ since the sketching matrix is a subspace embedding of $\mathbf{A}$. To see this, consider the smallest non-zero eigenvalue, $\lambda_1$:
\begin{align*}
\mathbf{A}v &= \lambda_1 v\\
|\mathbf{SA}v| &= (1+\epsilon)|\mathbf{A}v|_2 \\
&=(1+\epsilon)\lambda_1
\end{align*}
This agrees with the experimental results on the MNIST dataset. With the modified leverage score approximation algorithm, we recover the leverage scores to the approximation factor as guaranteed, from Figure~\ref{f5} (right). Using these insights, we summarize the following modifications to the Approximate Leverage Score algorithm, in Algorithm~\ref{a3}.

\vspace{-4mm}

\subsection{Practical Implementation Remarks}
We tested different implementations of sketched leverage scores to compare theoretical guarantees and practical performance. The sketches tested included Subsampled Randomized Hadamard Transform, CountSketch and OSNAP. In addition, sketching for CountSketch and OSNAP was implemented in a distributed setting using the central coordinator model. 

We found that the SRHT requirement of having rows as a power of 2 was impractical. The memory and timing overheads required to satisfy that requirement is prohibitive. OSNAP is more useful for real world data as it requires less rows. Most curated datasets do not have enough rows ($n$) so that CountSketch requirements can be satisfied.

\begin{figure}[!htb]
\centering
\includegraphics[scale=0.4]{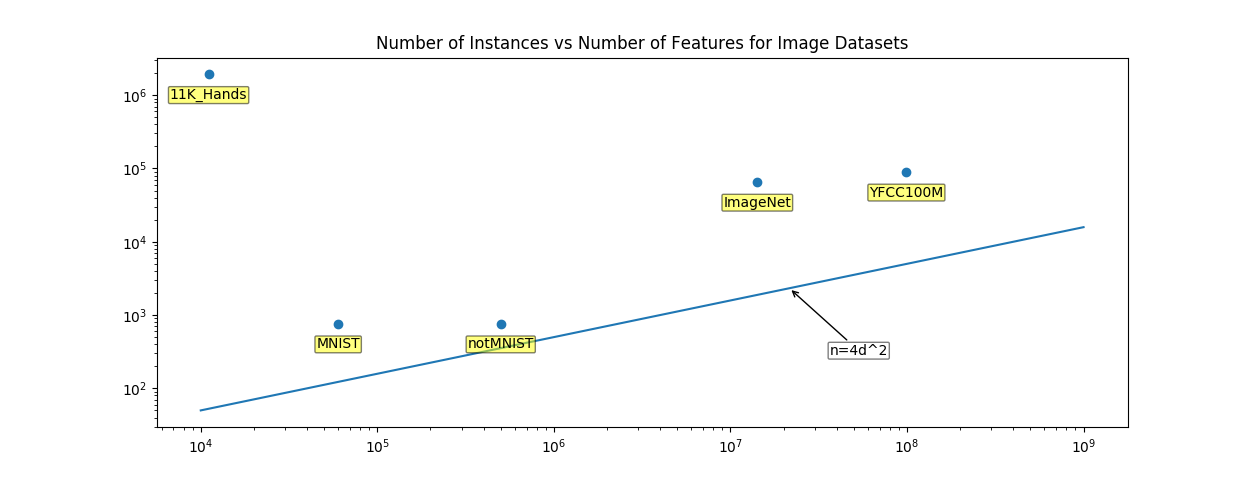}
\vspace{-8mm}
\caption{Popular image datasets for deep learning tasks. All of these datasets do not satisfy the $n \ge 4d^2$ requirement for CountSketch. The situation is worse when temporal data (i.e. text, videos, time series) is concerned since $d$ grows in both the feature and temporal dimensions.}
\end{figure}
\vspace{-4mm}

We found that the quality of the sketch is highly dependent on the column rank of the data. If the column rank of the data is less than the column dimension, then the approximation error can be unbounded. Requiring a full rank matrix for sketching as described in \citep{Clarkson:2017:LAR:3038256.3019134} is a hard requirement. 

Lastly, we found that small singular values in real data can corrupt the approximate leverage scores returned by sketching. This happens often as real data have high rank noise. We proposed a new algorithm to compute approximate leverage scores by first truncating the singular values before computing an approximate orthonormal basis. Experimentally, this modification recovers the approximate leverage scores to within the approximation guarantees given by sketching.

\section{\ourl}
\label{s4}
In this section, we describe an application of sketched leverage scores for training deep neural networks. The optimization of deep neural networks is largely based on stochastic gradient methods. Given the immense applications of deep neural networks, the empirical convergence and training speed of these networks and stochastic gradient methods have become important research areas. 

We focus on curriculum learning, a research area which aims to determine optimal orderings of training data to improve convergence and performance of neural networks~\citep{Bengio:2009:CL:1553374.1553380}. Curriculum Learning has traditionally focused on either manually determining a training order~\citep{Bengio:2009:CL:1553374.1553380,spitkovsky2010,NIPS2011_4466} or using auxiliary networks to jointly optimize for the optimal training order~\citep{NIPS2010_3923,Jiang:2015:SCL:2886521.2886696}. However, neither manual annotations nor costly overhead training times are ideal for training large neural networks on huge datasets. 

\noindent \begin{algorithm}[!htb]
\SetAlgoLined
\SetKwInOut{Input}{Input}\SetKwInOut{Output}{Output}
\caption{\ourl}
\Input{Given $n$ data points each with $d$ features}
1. Stack data points to form a $n\times d$ matrix $\mathbf{A}$

2. Compute leverage scores, $L=$\texttt{ApproxLeverageScoreWithTruncation}($\mathbf{A}$) 

3. Form the sampling probability $P_i$ for each data point from the leverage scores, $P = L_i/\Sigma_i L_i$
 
\For{each training epoch}{

4. Order the data points using sampling policy $\pi$ on distribution $P$.
    
5. Send the mini-batches for training in the given order.
}
\label{a4}
\end{algorithm}

To address the shortcomings in Curriculum Learning, we present \ourl \ (\ours), a novel technique for determining the ordering of data in the training of neural networks. \ours \ does not rely on manual annotations and is faster than policy gradient sampling methods in training auxiliary networks. 

\ours \ is shown in Algorithm~\ref{a4} and is based on approximating the leverage scores using matrix sketching techniques, so as to cater to the size of large datasets. These computed leverage scores provide a flexible and efficient method to determine the optimal ordering of training data without manual intervention or annotations. In particular, leverage scores can be used to estimate the importance of a training example. Larger leverage scores correspond to training examples that are more important. The sampling policy $\pi$ allows flexible design of the curriculum: leverage scores can be used independently or combined with an existing curriculum based on prior knowledge of the task. Furthermore, the leverage scores can be computed on either raw features or from semantic latent spaces using a pretrained network. 

\section{Experiments}
\label{s5}
\subsection{Datasets}

We perform extensive evaluations across 3 datasets from 3 different domains: CV (Computer Vision), NLP (Natural Language Processing) and multimodal machine learning.

\noindent \textbf{MNIST}~\citep{lecun-mnisthandwrittendigit-2010} is a collection of handwritten digits each labeled from 0 through 9. Results are reported in 10 class classification accuracy.

\noindent \textbf{SST}~\citep{socher2013recursive} is a collection of sentences from movie reviews each annotated with sentiment in the range [-2,2]. Results are reported in 5 class classification accuracy.

\noindent \textbf{CMU-MOSI}~\citep{zadeh2016multimodal} is a collection of 2199 opinion video clips. Each opinion video is annotated with sentiment in the range [-3,3]. Consistent with previous work~\citep{seq2seq}, Mean Absolute Error (MAE) is used as the loss function and results are reported in binary accuracy (A$^2$), F1-score (F1), MAE and Pearson correlation ($r$).

\subsection{Experimental Setup}
\label{orders}
Given these leverage scores, we use ideas from curriculum learning to generate 3 different orderings of the data based on several different sampling policies $\pi$: (1) \textbf{dec} indicates that the training data is ordered based on strictly decreasing leverage scores. As a result, the most important and diverse training points are seen first. (2) \textbf{dec, sampling with replacement} (\textbf{dec, swr}) indicates that the training data is ordered based on sampling with replacement from a discrete distribution defined by the leverage scores. As a result, the most important and diverse training points are seen first, but with randomness introduced into the order so that the order of training labels are less correlated. Note that sampling with replacement might duplicate datapoints with very high leverage scores and ignore datapoints with very low leverage scores. (3) \textbf{dec, sampling without replacement} (\textbf{dec, swor}) is the same but sampling without replacement. Additionally, we compare to the (4) \textbf{shuffle} baseline where models are trained on shuffled training data.





\subsection{Models for MNIST dataset}

We implement the following five models: \textbf{LR} is a Logistic Regression classifier \citep{Collins:2002:LRA:599615.599689}, \textbf{NN Small} is a small neural network, \textbf{NN Large} is a large neural network, \textbf{CNN Small} is a small convolutional neural network \citep{NIPS2012_4824}, \textbf{CNN Large} is a large convolutional neural network. Sketched leverage scores are computed for the $28 \times 28 = 784$ dimensional images.

\subsection{Models for SST dataset}

We use GloVe 300 dimensional word embeddings \citep{pennington2014glove} to convert words into vectors. A sequence length of 56 is selected and shorter sequence are zero-padded on the left. Longer sequences are truncated. We implement the following five models: \textbf{LR} is a Logistic Regression classifier, \textbf{DAN Small} is a small non-linear deep averaging network \citep{iyyer2015deep} that performs averaging of temporal features, \textbf{DAN Large} is a large non-linear deep averaging network, \textbf{LSTM Small} is a small Long Short Term Memory network~\citep{Hochreiter:1997:LSM:1246443.1246450}, \textbf{LSTM Large} is a large 2 layer Stacked Bidirectional Long Short Term Memory network~\citep{6638947}. Sketched leverage scores are computed for the $ 56 \times 300 = 16800$ dimensional sequences.

\subsection{Models for CMU-MOSI dataset}

We process the features and align the data using the same methods as~\citep{factorized}. Early Fusion (concatenation of language, visual and acoustic modalities)~\citep{Chen:2017:MSA:3136755.3136801} is performed at each time step to form a temporal sequence of length 20. The same five models for the SST dataset are used here. Sketched leverage scores are computed for the $ 20 \times 325 = 9750$ dimensional sequences.


Throughout the comparisons across ordering methods, all other hyperparameters (e.g. network size, learning rates, batchsize, number of epochs) are kept constant. Different runs use different settings of optimizers, learning rates and batchsizes. All hyparameter details and model configurations can be found in the Table captions and supplementary material.

\subsection{\ourl \ Results}

Table~\ref{mnist0} shows the accuracies for image classification on the MNIST dataset. Table~\ref{sst1} shows the accuracies for sentiment classification on the SST dataset. Table~\ref{mosi3} show the results for multimodal sentiment analysis on the CMU-MOSI dataset over 2 different runs using an early fusion method. For results on the CMU-MOSI dataset using late fusion methods and convergence graphs, please refer to the supplementary material.

\begin{table}[ht!]
\fontsize{7}{10}\selectfont
\centering
\setlength\tabcolsep{9.35pt}
\begin{tabular}{l : c c c c c}
\Xhline{3\arrayrulewidth}
Task & \multicolumn{5}{c}{\textbf{MNIST} 10 Class Image Classification Accuracy (\%)} \\
Method       & LR & NN Small & NN Large & CNN Small & CNN Large  \\ 
\Xhline{0.5\arrayrulewidth}
shuffle&92.84&98.50&98.28&98.98&99.34\\
dec&89.09&98.43&98.34&\textbf{99.01}&98.99\\
dec, swr&92.70&98.42&98.46&\textbf{99.01}&99.35\\
dec, swor&\textbf{92.88}&\textbf{98.55}&\textbf{98.57}&\textbf{99.01}&\textbf{99.39}\\
\Xhline{3\arrayrulewidth}
\end{tabular}
\caption{Results on MNIST dataset. 50 epochs, batchsize 256, Adam optimizer learning rate 0.001.}
\label{mnist0}
\end{table}
\vspace{-8mm}
\begin{table}[ht!]
\fontsize{7}{10}\selectfont
\centering
\setlength\tabcolsep{9.35pt}
\begin{tabular}{l : c c c c c}
\Xhline{3\arrayrulewidth}
Task & \multicolumn{5}{c}{\textbf{SST} 5 Class Sentiment Classification Accuracy (\%)} \\
Method       & LR & DAN Small & DAN Large & LSTM Small & LSTM Large  \\ 
\Xhline{0.5\arrayrulewidth}
shuffle&42.22&39.68&40.27&{42.35}&41.67\\
dec&\textbf{42.94}&39.86&\textbf{42.76}&{42.35}&41.31\\
dec, samp w rep&41.22&\textbf{40.72}&{40.72}&\textbf{42.44}&\textbf{43.71}\\
dec, samp wo rep&42.26&39.41&40.14&41.27&41.36\\
\Xhline{3\arrayrulewidth}
\end{tabular}
\caption{Results on SST dataset. Epochs determined by validation set, batchsize 256, Adam optimizer, learning rate 0.005.}
\label{sst1}
\end{table}
\vspace{-8mm}
\begin{table}[ht!]
\fontsize{7}{10}\selectfont
\centering
\setlength\tabcolsep{1.0pt}
\begin{tabular}{l : c c c c : c c c c : c c c c : c c c c : c c c c}
\Xhline{3\arrayrulewidth}
Task & \multicolumn{20}{c}{\textbf{CMU-MOSI} Sentiment Analysis} \\
Method       & \multicolumn{4}{c:}{LR} & \multicolumn{4}{c:}{DAN Small} & \multicolumn{4}{c:}{DAN Large} & \multicolumn{4}{c:}{LSTM Small} & \multicolumn{4}{c}{LSTM Large} \\ 
Metric & A$^{2}$ & F1 & MAE & $r$ & A$^{2}$ & F1 & MAE & $r$ & A$^{2}$ & F1 & MAE & $r$ & A$^{2}$ & F1 & MAE & $r$ & A$^{2}$ & F1 & MAE & $r$ \\
\Xhline{0.5\arrayrulewidth}
shuffle &56.7&52.4&1.367&0.373&60.3&58.5&1.288&0.434&61.2&59.9&1.314&0.438&73.9&74.0&1.068&0.624&73.3&73.3&1.067&0.604\\
dec&\textbf{59.6}&\textbf{57.1}&\textbf{1.353}&\textbf{0.392}&\textbf{63.0}&\textbf{61.3}&1.276&\textbf{0.460}&59.0&56.0&1.365&0.415&73.5&73.5&1.073&\textbf{0.626}&73.5&73.4&\textbf{1.038}&\textbf{0.621}\\
dec, swr &56.7&53.1&1.356&0.389&61.4&60.3&\textbf{1.274}&0.440&60.1&58.0&1.336&\textbf{0.413}&\textbf{74.6}&\textbf{74.7}&\textbf{1.061}&0.620&\textbf{74.1}&\textbf{73.9}&1.043&0.612\\
dec, swor &58.5&55.8&\textbf{1.353}&0.360&59.6&57.0&1.315&0.436&\textbf{64.3}&\textbf{64.3}&\textbf{1.271}&0.432&73.5&73.5&1.068&0.623&72.9&72.6&1.057&0.600\\
\Xhline{3\arrayrulewidth}
\end{tabular}
\caption{Results on CMU-MOSI dataset. Epochs determined by validation set, batchsize 128, Adam optimizer, learning rate 0.001.}
\label{mosi3}
\end{table}
\vspace{-8mm}
\section{Observations}
\label{s6}

Given these empirical results on convergence rates and final accuracies, we make the following observations: (1) changing the sequence which the training data is presented during training does affect the final testing accuracy. This shows that the conventional setup of uniform sampling during training can be improved upon. 
(2) on MNIST, decreasing order without sampling improves performance. (3) on SST, decreasing order with replacement improves performance. (4) on CMU-MOSI, sampling in a decreasing order improves accuracies.
\section{Related Works}
\label{s7}
\subsection{Leverage Scores}
~\citep{Clarkson:2017:LAR:3038256.3019134} developed and gave an overview of sketching techniques with provable guarantees. ~\citep{DBLP:journals/corr/BalcanLSW015} proposed distributed leverage scores based on sketching methods to speed up computation. Previous work proposed using leverage score to select data points in an active learning setup~\citep{DBLP:journals/corr/OrhanT15}. ~\citep{DBLP:conf/kdd/DahiyaKW18} did an empirical evaluation of sketching for on general algorithms rather than a treatment of SVD problems.\ours \ differs in the efficient computation of sketched leverage scores and more extensive evaluations on multiple datasets and models spanning CV, NLP and multimodal tasks.

\subsection{Curriculum Learning}

Curriculum learning studies how the order of training data affects model performance. Proposed methods include using clearer examples at the beginning of training and slowly introducing difficult examples~\citep{Bengio:2009:CL:1553374.1553380}. The curriculum is often obtained via heuristics unique to individual problems. For example, in the task of classifying geometric shapes, the curriculum was derived by an increasing variability in shapes~\citep{Bengio:2009:CL:1553374.1553380}. Other works have proposed using the length of a sentence as a curriculum for learning grammar induction~\citep{spitkovsky2010} or determining a curriculum for robot learning using human participants~\citep{NIPS2011_4466}. \ours \ is a flexible complement to these curriculum learning techniques by automatically deriving a ranking of the importance of each data point without the need for manual intervention or annotations.

More recently, proposed algorithms embed the curriculum design into the learning objective and jointly optimize the learning objective together with the curriculum~\citep{NIPS2010_3923,Jiang:2015:SCL:2886521.2886696}. Other approaches propose algorithms for automatically selecting the training path by reducing it to a multi-armed bandit problem~\citep{DBLP:journals/corr/GravesBMMK17} or a Markov decision process~\citep{DBLP:journals/corr/FanTQBL17}. These techniques often train an auxiliary network using REINFORCE~\citep{Williams:1992:SSG:139611.139614} algorithm or other policy gradient methods~\citep{DBLP:journals/corr/FanTQBL17} to learn the curriculum. These methods involve multiple rounds of sampling and further slows down the runtime. \ours \ computes leverage scores via an efficient sketching approach to minimize the overhead.
\section{Conclusion}
\label{s8}

In conclusion, we provide algorithmic implementations for sketched SVD problems on real-world, large-scale datasets, and more importantly, we provide a comprehensive empirical evaluation of these algorithms and provide guidelines on how to ensure accurate deployment to real-world data. As an application of sketched SVD, we presented \ourl \ (\ours), a novel technique for determining the ordering of data in the training of neural networks. \ours \ computes leverage scores efficiently via sketching, and these leverage scores provide a fast method to determine better orderings of training data. The strength of \ours \ is justified by an extensive set of experiments across CV, NLP and multi-modal tasks and datasets. Our method shows improvements in convergence and results. We believe that our method will inspire more research in using sketched leverage scores to analyze intermediate layers of neural networks, as well as to further understand the importance of each data point for online learning and never-ending learning.

\bibliography{citations}

\end{document}